\documentclass[runningheads]{llncs}
\usepackage{graphicx}

\usepackage{amsmath,amssymb,amsthm,amsfonts}
\usepackage{bm,bbm}
\usepackage{acronym}
\usepackage{enumitem}
\usepackage{balance}
\usepackage{xspace}
\usepackage[skip=3pt]{subcaption}
\usepackage[skip=3pt]{caption}
\usepackage{url}

\usepackage{marvosym}
\usepackage{mathrsfs}
\usepackage{arydshln}
\usepackage{array}
\usepackage{colortbl}
\usepackage{pifont}

\newcommand{\xmark}{\ding{55}}

\newcommand{\tlv}{{TLV}\xspace}
\newcommand{\model}{{STLV-Align}\xspace}

\usepackage{makecell}
\usepackage{multirow}
\usepackage{booktabs}

\definecolor{myblue}{RGB}{0,105,252}
\definecolor{myred}{RGB}{247,96,102}
\definecolor{mygray}{gray}{0.4}
\definecolor{red}{RGB}{139, 0, 0}
\definecolor{green}{RGB}{0, 100, 0}
\definecolor{gold}{RGB}{255, 125, 0}

\makeatletter
\DeclareRobustCommand\onedot{\futurelet\@let@token\@onedot}
\def\@onedot{\ifx\@let@token.\else.\null\fi\xspace}

\def\ie{\emph{i.e}\onedot}

\def\etc{\emph{etc}\onedot}

\makeatother

\acrodef{nlp}[NLP]{natural language processing}
\acrodef{plm}[PLM]{pretrained language model}
\acrodef{sota}[SOTA]{state-of-the-art}
\acrodef{bs}[BS]{Beam Search}
\acrodef{mhs}[MHS]{Metropolis-Hastings Sampling}
\acrodef{hs}[HS]{Hybrid Search}
\acrodef{uas}[UAS]{unlabeled attachment score}
\acrodef{dda}[DDA]{Directed Dependency Accuracy}
\acrodef{sota}[SOTA]{state-of-the-art}
\acrodef{pos}[POS]{part-of-speech}
\acrodef{asr}[ASR]{attacking success rate}
\acrodef{ppl}[PPL]{Perplexity score}
\acrodef{cqr}[CQR]{Conversational Question Reformulation}
\acrodef{cqa}[CQA]{Conversational Question Answering}
\acrodef{mcqr}[MTCQR]{Multi-Topic Conversational Question Reformulation}
\acrodef{amt}[AMT]{Amazon Mechanical Turk}
\acrodef{mtcl}[MTCL]{Multi-Topic Contrastive Learning}

\def\vx{{\bm{x}}}

\begin{document}
\title{Towards Comprehensive Multimodal Perception: Introducing the Touch-Language-Vision Dataset}
%
%
\author{Ning Cheng\inst{1} \and
You Li\inst{1}  \and
Jing Gao\inst{1} \and 
Bin Fang\inst{2} \and
Jinan Xu\inst{1} \and
Wenjuan Han \inst{1}
}
\authorrunning{Cheng et al.}
%
\institute{Beijing Key Lab of Traffic Data Analysis and Mining, Beijing Jiaotong University, Beijing, China \and
School of Artificial Intelligence, Beijing University of Posts and Telecommunications, Beijing,
China
}

%
\maketitle              
\begin{abstract}
  Tactility provides crucial support and enhancement for the perception and interaction capabilities of both humans and robots. Nevertheless, the multimodal research related to touch primarily focuses on visual and tactile modalities, with limited exploration in the domain of language. Beyond vocabulary, sentence-level descriptions contain richer semantics. Based on this, we construct a touch-language-vision dataset named \tlv (Touch-Language-Vision) by human-machine cascade collaboration, featuring sentence-level descriptions for multimode alignment. The new dataset is used to fine-tune our proposed lightweight training framework, \model (Synergistic Touch-Language-Vision Alignment), achieving effective semantic alignment with minimal parameter adjustments (1\%). Project Page: \url{https://xiaoen0.github.io/touch.page/}. 

\end{abstract}

\section{Introduction}
\label{sec:intro}

Tactile perception occupies a distinctive and pivotal role within the human sensory system, constituting a fundamental basis for our cognitive comprehension of the environment, coexisting harmoniously with other sensory modalities, such as vision and audition. Tactility allows us to perceive the texture, temperature, and hardness of objects \etc, and enables us to explore environments and perform intricate tasks, such as grasping and manipulating. The significance of touch is evident not only in humans \cite{dahiya2009tactile,johansson2009coding} but also in robotic applications \cite{hansen2022visuotactile,qi2023general}, where the acquisition and processing of tactile information are crucial for enhancing the perceptual capabilities and interaction efficiency of these applications.

Despite the undeniable significance of touch, tactile-related multimodel research predominantly focuses on the visual and tactile modalities \cite{dave2024multimodal,kerr2022self,yang2023generating}, with limited exploration in the domain of language. While there are some works related to language, they remain primarily at the lexical level, serving as labels for classification purposes \cite{gao2020supervised,yang2022touch,yuan2017connecting}. This arises from the heightened challenges associated with annotating lengthier texts, including intricate narratives and elevated expenses. 

Continuous innovation in image-to-text models \cite{achiam2023gpt,team2023gemini} enables the generation of fluent text from prompts and images, thereby offering opportunities for tactile annotation with longer texts. In this work, we introduce a tactile-related multimodel dataset, named \tlv (Touch-Language-Vision), through human-machine cascade collaborative annotation. \tlv incorporates three modalities: touch, language, and vision, with pairwise correspondence between any two modalities, aiming to strengthen alignment between touch and language. Compared to a set of vocabularies (\ie, lexical-level descriptions), the descriptions in \tlv are at the sentence level, capable of conveying more rich and more complete semantic information. 

To assess \tlv's efficacy, we employ it as the training dataset and present a lightweight unsupervised training method, \model (Synergistic Touch-Language-Vision
Alignment). This method maps all modalities to a
shared embedding space, enabling effective semantic alignment. To improve training efficiency,
we employ Low-Rank Adaptation (LoRA) \cite{hu2021lora} for fine-tuning and only 1\% of the parameters are adjusted. Subsequently, we evaluate the performance of \model on various tactile classification tasks using a cross-domain dataset. Experimental results demonstrate the potential of the \tlv dataset. 
This paper presents the following contributions:
\begin{itemize}
\setlength{\itemsep}{0pt}
\setlength{\parsep}{0pt}
\setlength{\parskip}{0pt}
  \item Introducing \tlv, a new touch-language-vision
  dataset with sentence-level descriptions annotated by human-machine cascade collaboration, Addressing the challenge of tactile annotation for longer texts.
  \item Proposing \model, a lightweight joint pretraining framework characterized by independence from labeled data, the utilization of a smaller dataset, the adjustment of model parameters, and acceptable performance.  
  \item Validating the effectiveness of our dataset and method and providing direction for further optimization on tactile-related tasks.
\end{itemize}

\section{Related Work}

\subsection{Tactile Perception}
Extracting and leveraging tactile information, encompassing surface texture, elasticity, and temperature, holds substantial promise for advancements in both robotics and AI research \cite{cui2020self,fazeli2019see,lin2019learning}. Current tactile sensors primarily rely on vision, employing a camera and illumination system to record deformations in a curved elastomeric gel. This structure has given rise to diverse perception systems, including GelSight \cite{calandra2017feeling,dong2017improved,gomes2021generation,johnson2011microgeometry,li2019connecting,si2022taxim,yang2022touch,yuan2017shape}, DIGIT \cite{kerr2022self,lambeta2020digit,suresh2023midastouch}, and GelSlim \cite{gao2023objectfolder}. These systems aim to comprehensively record high-resolution, detailed tactile information. Among them, GelSight stands out as one of the most widely used tactile perception systems, offering elaborate capture of depth, shear, and surface orientation. This work leverages sensor images primarily from GelSight.

\subsection{Tactile Datasets}
A significant challenge in learning from the tactile modality lies in the substantial human effort and time required to construct high-quality datasets. Despite this hurdle, the research community's continuous efforts have yielded several publicly available datasets:  Objectfolder 2.0 \cite{gao2021objectfolder} (featuring 1,000 implicitly represented objects generated through simulation), SSVTP \cite{kerr2022self} (containing 4.5K spatially aligned image-tactile pairs acquired using DIGIT), the Feeling of Success \cite{calandra2017feeling} (employing a two-finger gripper and GelSight sensor), Touch and Go 
 \cite{yang2022touch} (a high-quality, in-the-wild dataset encompassing diverse categories and quantitative visuo-tactile pairs) and VisGel \cite{li2019connecting} (a dataset comprising over 12K touch instances and 3 million vision-touch frames). However, these datasets are in the absence of rich textual descriptions, hindering their potential for realizing higher-level cross-modal alignment. This work addresses this limitation by incorporating detailed and qualitative captions, fostering a more comprehensive and advanced cross-modal understanding.

\subsection{Multimodal Alignment}
Effectively aligning semantics from diverse modalities is fundamental and crucial in multimodal research, yet it had been challenging to construct a high-dimensional joint embedding space incorporating features of different modalities. CLIP \cite{radford2021learning} achieved remarkable performance and generalization ability through self-supervised contrastive pretraining on a massive dataset of 400 million image-text pairs scraped from the internet. Subsequent works such as ALIGN \cite{jia2021scaling}, Flamingo \cite{alayrac2022flamingo}, Open-CLIP \cite{schuhmann2022laionb} have further advanced the field towards more robust and accurate alignment. Beyond vision and language modalities, significant research efforts aim to bridge the gap between even more diverse modalities, including 3D points \cite{liu2024openshape,xue2023ulip}, audio \cite{guzhov2022audioclip,chen2023iquery}. ImageBind \cite{girdhar2023imagebind} significantly extended the joint embedding space to encompass six distinct modalities through image-centered contrastive learning, further promoting comprehensive cross-modal understanding. Along this line, LanguageBind \cite{zhu2023languagebind} proposed a language-centered alignment strategy to fully leverage the rich semantic information within the text, achieving significant performance improvements. This work builds upon these advancements by further expanding cross-modal alignment to concurrently include touch alongside other modalities.

\section{\tlv Dataset}
\label{sec:dataset}

\begin{figure*}[ht]
    \small
        \centering       
        \includegraphics[width=0.95\linewidth]{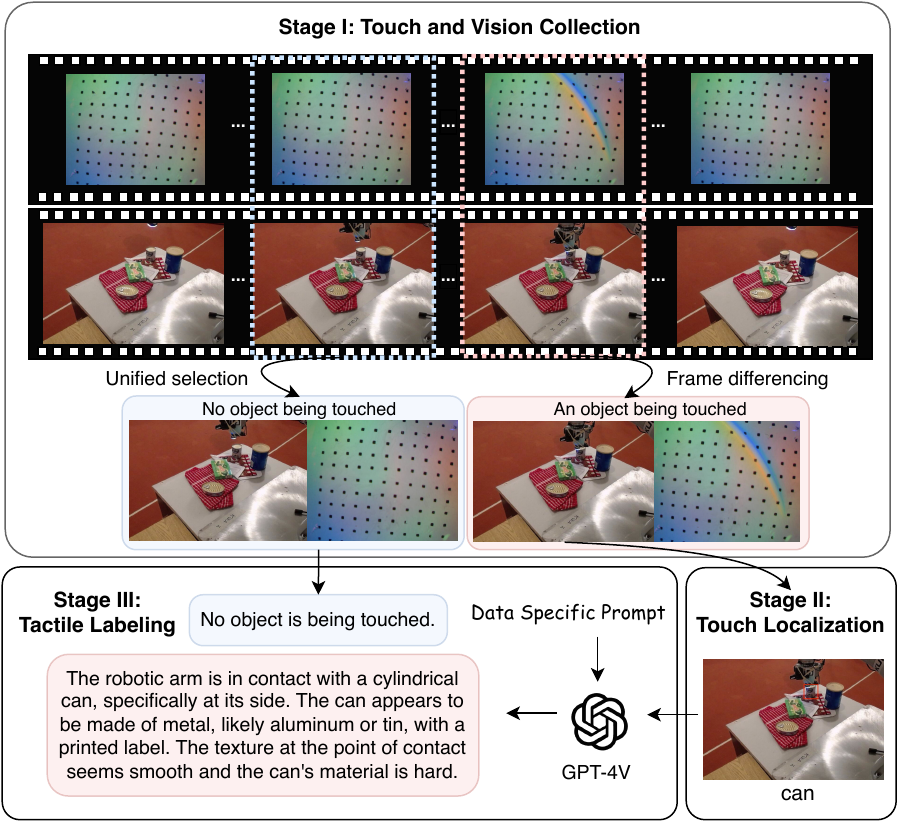}

        \caption{Construction process of the TLV dataset.}       
        \label{fig:tlv_dataset}
    \end{figure*}

The \tlv dataset aims to associate tactile and visual perceptions with sentence-level descriptions for multimodal alignment.
As shown in Figure \ref{fig:tlv_dataset}, the construction process of \tlv consists of three stages: touch and vision collection (Sec. \ref{sec:data_s_1}), touch localization (Sec. \ref{sec:data_s_2}), and tactile labeling (Sec. \ref{sec:data_s_3}).

\subsection{Stage I: Touch and Vision Collection}
\label{sec:data_s_1}
We collect paired tactile and visual observations from VisGel \cite{li2019connecting}, a large vision-touch dataset collected by a video camera, and a tactile sensor called GelSight \cite{johnson2011microgeometry}.
VisGel captured synchronized videos of the scenes where the robotic arm touched the objects and recorded timestamps to synchronize visual and tactile images. Among the synchronized videos captured, 10,000 videos were used to construct the training dataset. 
We utilize the synchronized visual and tactile images from these 10,000 synchronized videos for touch and vision collection. 

From the visual videos, we observe that the first frame depicts the starting state of the robotic arm when it is away from objects. As time progresses, the arm gradually approaches an object until it makes contact, remains in contact for a period, and then slowly withdraws. 
Based on the above observations, for each pair of synchronized videos, we select two sets of synchronized visual and tactile frames: one set depicting an object being touched, and the other set showing no object being touched.
To obtain frames where an object is being touched, we use the first frame as the background and apply frame differencing \cite{zaki2011moving}. The frame with the maximum difference from the background is selected as the frame where the object is being touched. Through observation, we uniformly select the 40th frame as the frame where no object is being touched.

\subsection{Stage II: Touch Localization}
\label{sec:data_s_2}
The two modalities of touch and vision can be regarded as different views containing the same semantics. From this standpoint, we recruit participants to label the object being touched in visual images from Stage I. For visual images where no object is touched, we do not consider them. This serves as preliminary touch localization, preparing for the next step of tactile labeling using GPT-4V \cite{achiam2023gpt}. touch localization comprises two parts: highlighting the touched object with a red box in visual images and providing a name for the enclosed object. The labeling of object name is open-ended, and we do not provide a predefined set of candidate object names. In the process of object labeling, we found that due to issues during the collection of the original dataset (\ie VisGel), certain data could not be annotated. For example, instances where the touched object is occluded or the entire video does not involve interaction with any object. We have filtered out such data. 

\subsection{Stage III: Tactile Labeling}
\label{sec:data_s_3}
From the perspective of containing identical semantics in both touch and vision, we utilize GPT-4V for the annotation of texts. For each visual image with the highlighted box from Stage II, we employ a thoughtfully designed, data-specific prompt. This prompt instructs GPT-4V to generate detailed descriptions, taking into account factors such as the name of the touched object, the specific location of the contact, the material composition at the point of contact, and the texture characteristics and softness/hardness of the touched area. For visual images where no object is touched, we refrain from using GPT-4V for annotation and instead provide a uniform description: \textit{No object is being touched.}.

\subsection{Dataset Statistics}
We have annotated text-based descriptions for 20,000 pairs of synchronized tactile and visual observations collected from VisGel, including 10,000 pairs with an object being touched and 10,000 pairs without an object being touched. For the cases where an object is touched, we filtered out data that cannot be annotated as mentioned in Stage II, resulting in the annotation of 9,834 instances. For the cases without an object being touched, we annotated all 10,000 instances. Thus, we ultimately obtained a total of 19,834 annotated data entries. To our knowledge, this is the first touch-language-vision dataset with sentence-level descriptions.

\section{Method}
\begin{figure}[ht]
    \small
        \centering       
        \includegraphics[width=0.6\linewidth]{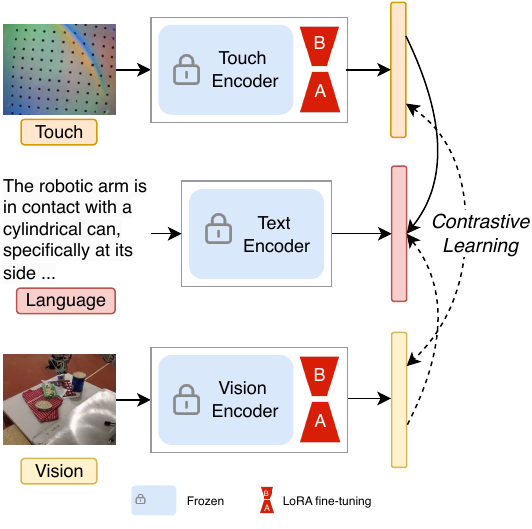}

        \caption{Overview of our lightweight joint training method.}       
        \label{fig:method}
    \end{figure}

We propose \model (Synergistic Touch-Language-Vision Alignment), an unsupervised and lightweight joint training method designed to leverage the TLV dataset we constructed. The visual observations in TLV can be considered as auxiliary information, assisting in learning the alignment between touch and language and enhancing the zero-shot classification ability of touch. The method primarily consists of three components: multi-modal encoder, LoRA fine-tuning, and joint training, as illustrated in Fig. \ref{fig:method}.

\subsection{Multi-modal Encoders}
\model involves three modalities: touch, language, and vision. We treat the touch modality as RGB images for processing. Therefore, for both touch and vision modalities, we use the Vision Transformer (\ie, ViT) \cite{dosovitskiy2020image} for encoding. The touch and vision encoders are instantiated as OpenCLIP vision encoders. For the text encoder, we instantiate it as an OpenCLIP text encoder.

\subsection{LoRA Fine-tuning}
Differing from the prior approach \cite{lei2023vit}, we do not utilize large-scale datasets for pre-training. Instead, we employ LoRA for lightweight fine-tuning on the TLV dataset.  For a modality-agnostic encoder $f(\cdot)$ with a weight matrix $W_0\in \mathbb{R}^{d\times k}$, we maintain the weight matrix $W_0$ frozen while learning a new weight matrix $BA$. The forward pass can be formalized as follows:
\begin{equation}
\label{eq:lora}
f({\displaystyle \vx}) = W_0 {\displaystyle \vx} + BA \boldsymbol{x}
\end{equation}
where $B\in \mathbb{R}^{d\times r}, A\in \mathbb{R}^{r\times k}$, with $r$ being the minimum of $d$ and $k$. 

\subsection{Joint Training}
Joint learning aims to align touch and language better. While learning the alignment between touch and language, we also acquire knowledge about the alignment between vision and language, as well as the alignment between touch and vision. The text encoder from OpenCLIP has demonstrated good generalization in text, so during the joint learning process, we freeze the text encoder and only update the touch encoder and vision encoder. The update to the vision encoder is made to assist the update to the touch encoder.
To ensure alignment across different modalities, we perform contrastive learning principles \cite{radford2021learning} for joint learning.
\begin{equation}\label{eq:loss}
\begin{aligned}
L_{T,L}=-\frac{1}{K} \sum_{i=1}^{K} \log \frac{\exp (x_i^{\top} y_i / \tau)}{\sum_{j=1}^{K} \exp (x_i^{\top} y_j / \tau)},    \\
L_{V,L}=-\frac{1}{K} \sum_{i=1}^{K} \log \frac{\exp (z_i^{\top} y_i / \tau)}{\sum_{j=1}^{K} \exp (z_i^{\top} y_j / \tau)}, \\
L_{T,V}=-\frac{1}{K} \sum_{i=1}^{K} \log \frac{\exp (x_i^{\top} z_i / \tau)}{\sum_{j=1}^{K} \exp (x_i^{\top} z_j / \tau)} \\
\end{aligned}
\end{equation}
where $x, y, z$ represent the observations of tactile, language, and visual modalities, respectively, and $\tau$ and $K$ are the scalar temperature and batch size. In practice, we
use a symmetric joint loss $(L_{T,L}+L_{L,T})+\alpha(L_{V,L}+L_{L,V})+\beta(L_{T,V}+L_{V,T})$.

\section{Experiments}

\subsection{Setup}
We evaluate our model and dataset on various tactile classification tasks, including material, hard/soft, and rough/smooth classification, using the Touch-and-Go dataset \cite{yang2022touch}. This means that zero-shot evaluation is conducted on a cross-domain dataset. \model is extended based on OpenCLIP-large \cite{ilharco_gabriel_2021_5143773} and fine-tuned on our \tlv dataset in an unsupervised and lightweight manner. Because the visual modality is considered as auxiliary information, both $\alpha$ and $\beta$ are set to 0.1 in the symmetric joint loss. We use accuracy as the metric.

\subsection{Results and Analysis}
We contrast our model with ViT-LENS-2 \cite{lei2023vit}, a state-of-the-art multi-model that excels in zero-shot performance on tactile tasks. The comparative results of different models can be found in Table \ref{tab:main-result}. While the accuracy of \model may be not optimal, exhibits an 8.3\% improvement in material classification compared to our foundation, OpenCLIP. Especially significant is the marked improvement in both hard/soft and rough/smooth classifications, with \model's performance advancing by more than 30\%. Nevertheless, VIT-LENS-2 (I) demonstrated an improvement ranging from 7\% to 9\% compared to their foundation, ImageBind \cite{girdhar2023imagebind}. Despite VIT-LENS-2 (I+T) displaying a notable 41.6\% boost in material classification, there was a 6\% decline in rough/smooth classification. This reflects the effectiveness of the TLV dataset and the efficiency of \model in utilizing data. Certainly, we recognize certain performance limitations of \model, thus prompting us to analyze distinctions in training paradigm, \#training data, parameter tuning ratio, and cross-domain evaluation between \model and ViT-LENS-2, as illustrated in Table \ref{tab:comparison_approach}. It can be observed that \model is characterized by independence from labeled data, the utilization of a smaller dataset, a lightweight training approach, and evaluation across various data domains. This could make it more attractive for specific application scenarios.

\begin{table}
\vspace{-5mm}
\centering
\caption{Accuracy of different models on various tactile classifications. Those with performance improvement compared with their respective foundation within 30\% are marked as \textcolor{green}{green} and above 30\% are marked as \textcolor{myred}{red}. V-L-2: VIT-LENS-2; (I): Anchored by images; (I+T): Anchored by images and texts.} 
\label{tab:main-result}
\setlength\tabcolsep{9pt} 
\begin{tabular}{|l|c|ccc|}
\hline
\multirow{2}{*}{\textbf{Model}} & \multirow{2}{*}{\textbf{Size}} & \multicolumn{3}{c|}{\textbf{Touch and Go}}  \\ 
\cline{3-5}
& & Material     & Hard/Soft    & Rough/Smooth \\
\hline
ImageBind  & Base & 24.2 & 65.7 & 69.8            \\
V-L-2 (I)  & Base & 29.9 \small (\textcolor{green}{+5.7\%}) & 72.4 \small (\textcolor{green}{+6.7\%}) & 77.9  \small (\textcolor{green}{+8.1\%})       \\
V-L-2 (I)  & Large & 31.2 \small (\textcolor{green}{+7.0\%}) & 74.3 \small (\textcolor{green}{+8.6\%}) & 78.2 \small (\textcolor{green}{+8.4\%})         \\
V-L-2 (I+T) & Large & 65.8 \small (\textcolor{myred}{+41.6\%}) & 74.7 \small (\textcolor{green}{+9.0\%}) &  63.8 \small (\textcolor{green}{-6.0\%})         \\
\hline
OpenCLIP  & Large & 17.7 & 32.2 & 42.7           \\
\model  & Large & 26.0 \small (\textcolor{green}{+8.3\%}) & 65.1 \small (\textcolor{myred}{+32.9\%}) & 74.6 \small (\textcolor{myred}{+31.9\%})          \\
\hline
\end{tabular}

\vspace{-6mm}
\end{table}
 
\begin{table}
\vspace{-6mm}
\centering
\caption{Comparison of Ours and VIT-LENS-2 in training paradigm (\textbf{TP}), \#training data (\textbf{\#TD}), parameter tuning ratio (\textbf{PTR}), cross-domain evaluation (\textbf{CDE}).}
\label{tab:comparison_approach}
\setlength\tabcolsep{17pt} 
\begin{tabular}{|c|cccc|} 
\hline
\textbf{Model} &\textbf{TP} & \textbf{\#TD} & \textbf{PTR} & \textbf{CDE}      \\ 
\hline
VIT-LENS-2  & Supervised & 91,982 & 100\%  &   \xmark   \\
\model & Unsupervised & 19,843 & 1\%    &   \checkmark \\
\hline
\end{tabular}

\end{table}

\subsection{Ablation Study}

We conduct the ablation study in Table \ref{tab:ablation} to illustrate the impact of vision information. Simultaneously aligning both touch and text with visual information enhances tactile classification, yielding a positive overall effect. Conversely, aligning either touch or text with visual information has a detrimental effect.

\begin{table}[!ht]
\small
\vspace{-3mm}
\centering
\caption{Impact of vision information at different levels. -TV: Do not align touch with vision; -VL: Do not align language with vision; -(TV\&VL): Do not involve visual information.} 
\label{tab:ablation}
\setlength\tabcolsep{20pt} 
\begin{tabular}{|c|ccc|}
\hline
\multirow{2}{*}{\textbf{Model}} & \multicolumn{3}{c|}{\textbf{Touch and Go}}  \\ 
\cline{2-4}
& Material  & Hard/Soft    & Rough/Smooth           \\ 
\hline
\model  &  26.0  & \textbf{65.1}  & \textbf{74.6}        \\
-TV  &  27.8  & 52.8  &  52.7   \\
-VL &  26.5  & 55.3  &  49.1   \\
-(TV\&VL) &  \textbf{32.5}  & 56.5  &  56.6  \\
\hline
\end{tabular}

\vspace{-5mm}
\end{table}

\section{Conclusion}
In this work, we construct the first touch-language-vision dataset, \tlv, featuring sentence-level descriptions for multimodal alignment. To demonstrate the effectiveness of the \tlv dataset, we extended OpenCLIP and proposed \model, an unsupervised lightweight training approach. Preliminary experiments validate that the \tlv dataset facilitates better alignment between touch and language. The proposed method may apply to specific scenarios, but there is room for improvement in terms of performance, and further enhancements are needed. Additionally, we intend to extend the application of \tlv to more tasks to fully exploit its potential.

%
%
%
\bibliographystyle{splncs04}
\bibliography{custom}

\begin{thebibliography}{10}
\providecommand{\url}[1]{\texttt{#1}}
\providecommand{\urlprefix}{URL }
\providecommand{\doi}[1]{https://doi.org/#1}

\bibitem{achiam2023gpt}
Achiam, J., Adler, S., Agarwal, S., Ahmad, L., Akkaya, I., Aleman, F.L., Almeida, D., Altenschmidt, J., Altman, S., Anadkat, S., et~al.: Gpt-4 technical report. arXiv preprint arXiv:2303.08774  (2023)

\bibitem{alayrac2022flamingo}
Alayrac, J.B., Donahue, J., Luc, P., Miech, A., Barr, I., Hasson, Y., Lenc, K., Mensch, A., Millican, K., Reynolds, M., et~al.: Flamingo: a visual language model for few-shot learning. Advances in neural information processing systems  \textbf{35},  23716--23736 (2022)

\bibitem{calandra2017feeling}
Calandra, R., Owens, A., Upadhyaya, M., Yuan, W., Lin, J., Adelson, E.H., Levine, S.: The feeling of success: Does touch sensing help predict grasp outcomes? arXiv preprint arXiv:1710.05512  (2017)

\bibitem{chen2023iquery}
Chen, J., Zhang, R., Lian, D., Yang, J., Zeng, Z., Shi, J.: iquery: Instruments as queries for audio-visual sound separation. In: Proceedings of the IEEE/CVF Conference on Computer Vision and Pattern Recognition. pp. 14675--14686 (2023)

\bibitem{cui2020self}
Cui, S., Wang, R., Wei, J., Hu, J., Wang, S.: Self-attention based visual-tactile fusion learning for predicting grasp outcomes. IEEE Robotics and Automation Letters  \textbf{5}(4),  5827--5834 (2020)

\bibitem{dahiya2009tactile}
Dahiya, R.S., Metta, G., Valle, M., Sandini, G.: Tactile sensing—from humans to humanoids. IEEE transactions on robotics  \textbf{26}(1),  1--20 (2009)

\bibitem{dave2024multimodal}
Dave, V., Lygerakis, F., R{\"u}ckert, E.: Multimodal visual-tactile representation learning through self-supervised contrastive pre-training. In: Proceedings/IEEE International Conference on Robotics and Automation. Institute of Electrical and Electronics Engineers (2024)

\bibitem{dong2017improved}
Dong, S., Yuan, W., Adelson, E.H.: Improved gelsight tactile sensor for measuring geometry and slip. In: 2017 IEEE/RSJ International Conference on Intelligent Robots and Systems (IROS). pp. 137--144 (2017)

\bibitem{dosovitskiy2020image}
Dosovitskiy, A., Beyer, L., Kolesnikov, A., Weissenborn, D., Zhai, X., Unterthiner, T., Dehghani, M., Minderer, M., Heigold, G., Gelly, S., et~al.: An image is worth 16x16 words: Transformers for image recognition at scale. In: International Conference on Learning Representations (2020)

\bibitem{fazeli2019see}
Fazeli, N., Oller, M., Wu, J., Wu, Z., Tenenbaum, J.B., Rodriguez, A.: See, feel, act: Hierarchical learning for complex manipulation skills with multisensory fusion. Science Robotics  \textbf{4}(26),  eaav3123 (2019)

\bibitem{gao2020supervised}
Gao, R., Taunyazov, T., Lin, Z., Wu, Y.: Supervised autoencoder joint learning on heterogeneous tactile sensory data: Improving material classification performance. In: 2020 IEEE/RSJ International Conference on Intelligent Robots and Systems (IROS). pp. 10907--10913. IEEE (2020)

\bibitem{gao2021objectfolder}
Gao, R., Chang, Y.Y., Mall, S., Fei-Fei, L., Wu, J.: Objectfolder: A dataset of objects with implicit visual, auditory, and tactile representations. In: 5th Annual Conference on Robot Learning (2021)

\bibitem{gao2023objectfolder}
Gao, R., Dou, Y., Li, H., Agarwal, T., Bohg, J., Li, Y., Fei-Fei, L., Wu, J.: The objectfolder benchmark: Multisensory learning with neural and real objects. In: IEEE/CVF Conference on Computer Vision and Pattern Recognition (2023)

\bibitem{girdhar2023imagebind}
Girdhar, R., El-Nouby, A., Liu, Z., Singh, M., Alwala, K.V., Joulin, A., Misra, I.: Imagebind: One embedding space to bind them all. In: Proceedings of the IEEE/CVF Conference on Computer Vision and Pattern Recognition. pp. 15180--15190 (2023)

\bibitem{gomes2021generation}
Gomes, D.F., Paoletti, P., Luo, S.: Generation of gelsight tactile images for sim2real learning. IEEE Robotics and Automation Letters  \textbf{6}(2),  4177--4184 (2021)

\bibitem{guzhov2022audioclip}
Guzhov, A., Raue, F., Hees, J., Dengel, A.: Audioclip: Extending clip to image, text and audio. In: ICASSP 2022-2022 IEEE International Conference on Acoustics, Speech and Signal Processing (ICASSP). pp. 976--980. IEEE (2022)

\bibitem{hansen2022visuotactile}
Hansen, J., Hogan, F., Rivkin, D., Meger, D., Jenkin, M., Dudek, G.: Visuotactile-rl: Learning multimodal manipulation policies with deep reinforcement learning. In: 2022 International Conference on Robotics and Automation (ICRA). pp. 8298--8304. IEEE (2022)

\bibitem{hu2021lora}
Hu, E.J., Wallis, P., Allen-Zhu, Z., Li, Y., Wang, S., Wang, L., Chen, W., et~al.: Lora: Low-rank adaptation of large language models. In: International Conference on Learning Representations (2021)

\bibitem{ilharco_gabriel_2021_5143773}
Ilharco, G., Wortsman, M., Wightman, R., Gordon, C., Carlini, N., Taori, R., Dave, A., Shankar, V., Namkoong, H., Miller, J., Hajishirzi, H., Farhadi, A., Schmidt, L.: Openclip (Jul 2021). \doi{10.5281/zenodo.5143773}, \url{https://doi.org/10.5281/zenodo.5143773}

\bibitem{jia2021scaling}
Jia, C., Yang, Y., Xia, Y., Chen, Y.T., Parekh, Z., Pham, H., Le, Q., Sung, Y.H., Li, Z., Duerig, T.: Scaling up visual and vision-language representation learning with noisy text supervision. In: International conference on machine learning. pp. 4904--4916. PMLR (2021)

\bibitem{johansson2009coding}
Johansson, R.S., Flanagan, J.R.: Coding and use of tactile signals from the fingertips in object manipulation tasks. Nature Reviews Neuroscience  \textbf{10}(5),  345--359 (2009)

\bibitem{johnson2011microgeometry}
Johnson, M.K., Cole, F., Raj, A., Adelson, E.H.: Microgeometry capture using an elastomeric sensor. ACM Transactions on Graphics (TOG)  \textbf{30}(4), ~1--8 (2011)

\bibitem{kerr2022self}
Kerr, J., Huang, H., Wilcox, A., Hoque, R., Ichnowski, J., Calandra, R., Goldberg, K.: Self-supervised visuo-tactile pretraining to locate and follow garment features. arXiv preprint arXiv:2209.13042  (2022)

\bibitem{lambeta2020digit}
Lambeta, M., Chou, P.W., Tian, S., Yang, B., Maloon, B., Most, V.R., Stroud, D., Santos, R., Byagowi, A., Kammerer, G., et~al.: Digit: A novel design for a low-cost compact high-resolution tactile sensor with application to in-hand manipulation. IEEE Robotics and Automation Letters  \textbf{5}(3),  3838--3845 (2020)

\bibitem{lei2023vit}
Lei, W., Ge, Y., Yi, K., Zhang, J., Gao, D., Sun, D., Ge, Y., Shan, Y., Shou, M.Z.: Vit-lens-2: Gateway to omni-modal intelligence. arXiv preprint arXiv:2311.16081  (2023)

\bibitem{li2019connecting}
Li, Y., Zhu, J.Y., Tedrake, R., Torralba, A.: Connecting touch and vision via cross-modal prediction. In: Proceedings of the IEEE/CVF Conference on Computer Vision and Pattern Recognition. pp. 10609--10618 (2019)

\bibitem{lin2019learning}
Lin, J., Calandra, R., Levine, S.: Learning to identify object instances by touch: Tactile recognition via multimodal matching. In: 2019 International Conference on Robotics and Automation (ICRA). pp. 3644--3650 (2019)

\bibitem{liu2024openshape}
Liu, M., Shi, R., Kuang, K., Zhu, Y., Li, X., Han, S., Cai, H., Porikli, F., Su, H.: Openshape: Scaling up 3d shape representation towards open-world understanding. Advances in Neural Information Processing Systems  \textbf{36} (2024)

\bibitem{qi2023general}
Qi, H., Yi, B., Suresh, S., Lambeta, M., Ma, Y., Calandra, R., Malik, J.: General in-hand object rotation with vision and touch. In: Conference on Robot Learning. pp. 2549--2564. PMLR (2023)

\bibitem{radford2021learning}
Radford, A., Kim, J.W., Hallacy, C., Ramesh, A., Goh, G., Agarwal, S., Sastry, G., Askell, A., Mishkin, P., Clark, J., et~al.: Learning transferable visual models from natural language supervision. In: International conference on machine learning. pp. 8748--8763. PMLR (2021)

\bibitem{schuhmann2022laionb}
Schuhmann, C., Beaumont, R., Vencu, R., Gordon, C.W., Wightman, R., Cherti, M., Coombes, T., Katta, A., Mullis, C., Wortsman, M., Schramowski, P., Kundurthy, S.R., Crowson, K., Schmidt, L., Kaczmarczyk, R., Jitsev, J.: {LAION}-5b: An open large-scale dataset for training next generation image-text models. In: Thirty-sixth Conference on Neural Information Processing Systems Datasets and Benchmarks Track (2022), \url{https://openreview.net/forum?id=M3Y74vmsMcY}

\bibitem{si2022taxim}
Si, Z., Yuan, W.: Taxim: An example-based simulation model for gelsight tactile sensors. IEEE Robotics and Automation Letters  \textbf{7}(2),  2361--2368 (2022)

\bibitem{suresh2023midastouch}
Suresh, S., Si, Z., Anderson, S., Kaess, M., Mukadam, M.: Midastouch: Monte-carlo inference over distributions across sliding touch. In: Conference on Robot Learning. pp. 319--331 (2023)

\bibitem{team2023gemini}
Team, G., Anil, R., Borgeaud, S., Wu, Y., Alayrac, J.B., Yu, J., Soricut, R., Schalkwyk, J., Dai, A.M., Hauth, A., et~al.: Gemini: a family of highly capable multimodal models. arXiv preprint arXiv:2312.11805  (2023)

\bibitem{xue2023ulip}
Xue, L., Gao, M., Xing, C., Mart{\'\i}n-Mart{\'\i}n, R., Wu, J., Xiong, C., Xu, R., Niebles, J.C., Savarese, S.: Ulip: Learning a unified representation of language, images, and point clouds for 3d understanding. In: Proceedings of the IEEE/CVF Conference on Computer Vision and Pattern Recognition. pp. 1179--1189 (2023)

\bibitem{yang2022touch}
Yang, F., Ma, C., Zhang, J., Zhu, J., Yuan, W., Owens, A.: Touch and go: Learning from human-collected vision and touch. Advances in Neural Information Processing Systems  \textbf{35},  8081--8103 (2022)

\bibitem{yang2023generating}
Yang, F., Zhang, J., Owens, A.: Generating visual scenes from touch. In: Proceedings of the IEEE/CVF International Conference on Computer Vision. pp. 22070--22080 (2023)

\bibitem{yuan2017connecting}
Yuan, W., Wang, S., Dong, S., Adelson, E.: Connecting look and feel: Associating the visual and tactile properties of physical materials. In: Proceedings of the IEEE conference on computer vision and pattern recognition. pp. 5580--5588 (2017)

\bibitem{yuan2017shape}
Yuan, W., Zhu, C., Owens, A., Srinivasan, M.A., Adelson, E.H.: Shape-independent hardness estimation using deep learning and a gelsight tactile sensor. In: 2017 IEEE International Conference on Robotics and Automation (ICRA). pp. 951--958 (2017)

\bibitem{zaki2011moving}
Zaki, W.M.D.W., Hussain, A., Hedayati, M.: Moving object detection using keypoints reference model. EURASIP J. Image Video Process.  \textbf{2011}(1), ~13 (2011)

\bibitem{zhu2023languagebind}
Zhu, B., Lin, B., Ning, M., Yan, Y., Cui, J., HongFa, W., Pang, Y., Jiang, W., Zhang, J., Li, Z., et~al.: Languagebind: Extending video-language pretraining to n-modality by language-based semantic alignment. In: The Twelfth International Conference on Learning Representations (2023)

\end{thebibliography}

\end{document}